\documentclass[12pt,a4paper]{article}

\usepackage{times} 
\usepackage[a4paper, lmargin=2cm, rmargin=2cm , tmargin=2.5cm , bmargin=2.25cm  ,footskip=1.25cm]{geometry} 
\usepackage{graphicx} 
\usepackage[square,sort,comma,numbers]{natbib} 
\usepackage{amsmath} 
\usepackage[hyphens]{url} 
\usepackage{soul}
\usepackage{fancyhdr} 
\usepackage[explicit]{titlesec} 
\usepackage{abstract} 
\usepackage{authblk} 
\usepackage{tabularx}
\usepackage{multirow} 
\usepackage{colortbl} 
\usepackage{booktabs}
\usepackage{array}
\usepackage[super]{nth} 
\usepackage{algorithm}
\usepackage[end]{algpseudocode}
\usepackage{amsfonts}

\titleformat{\section}{\normalfont\bfseries}{\thesection}{1em}{\MakeUppercase{#1}}  

\titleformat{\subsection}{\normalfont\bfseries\small}{\thesubsection}{1em}{{#1}} 

\titleformat{\subsubsection}{\normalfont\small}{\thesubsubsection}{1em}{{#1}} 

\algrenewcommand\algorithmicend{\textbf{end}}
\algrenewtext{EndFor}{\algorithmicend}

\makeatletter
\newcommand{\thickhline}{ \noalign {\ifnum 0=`} \fi \hrule height 1pt \futurelet \reserved@a \@xhline }
\newcommand{\morethickhline}{ \noalign {\ifnum 0=`} \fi \hrule height 2pt \futurelet \reserved@a \@xhline }
\newcolumntype{"}{@{\hskip\tabcolsep\vrule width 1pt\hskip\tabcolsep}}
\makeatother

\newlength{\Oldarrayrulewidth}

\begin{document}

\title{\normalsize\normalfont\bfseries \MakeUppercase {Reduced-Order-Model-Based Feedback Design for Thruster-Assisted Legged Locomotion} }
\date{}

\author[]{Pravin Dangol}
\author[]{Alireza Ramezani}
\affil[]{ \textit{Electrical and Computer Engineering, Northeastern University, Boston, USA}}

\renewcommand\Authands{, }
\renewcommand{\Authfont}{\normalsize\normalfont \bfseries}
\renewcommand{\Affilfont}{\normalsize\normalfont}
\renewcommand{\abstractnamefont}{\bfseries\normalsize\MakeUppercase} 

\maketitle

\begin{abstract}
Real-time constraint satisfaction for robots can be quite challenging due to the high computational complexity that arises when accounting for the system dynamics and environmental interactions, often requiring simplification in modelling that might not necessarily account for all performance criteria. 

We instead propose an optimization-free approach where reference trajectories are manipulated to satisfy constraints brought on by ground contact as well as those prescribed for states and inputs. Unintended changes to trajectories especially ones optimized to produce periodic gaits can adversely affect gait stability, however we will show our approach can still guarantee stability of a gait by employing the use of coaxial thrusters that are unique to our robot.
\end{abstract}

\thispagestyle{fancy}

\section{Introduction}

We have categorized constraint satisfaction in legged robots in three broad categories. Namely: i) trajectory optimization (TO), ii) optimization-based controls and iii) reference trajectory manipulation. i) The goal of the TO problem is to generate optimal trajectory which satisfy constraints on states, inputs and ground reaction forces (GRF) while ensuring that the trajectories lead to stable walking gaits. 

TO problems for legged robot are difficult to solve due to their nonlinear dynamics, high degrees of freedom and the hybrid nature of the system brought on by ground impact \cite{ramezani_atrias_2012,park_finite-state_2013,park_switching_2012,buss_preliminary_2014,ramezani_performance_2014}. Previous works such as \cite{posa2014direct}, \cite{hereid_3d_2016}, \cite{medeiros2020trajectory} have proposed methods to transcribe this as a non-linear programming (NLP) problem through direct collocation methods where polynomial splines are used to approximate the continuous dynamics and thus reducing computational complexity without needing to account for the actual dynamics. While \cite{hereid2015hybrid} has instead proposed utilizing multiple shooting method to break the original problem down into smaller steps without approximations.

In both cases, however, the dynamics of the system needs to be considered along with contact dynamics to generate the trajectories. To alleviate the need of explicitly defining ground contact dynamics \cite{pardo_hybrid_2017}, \cite{xin2020optimization} have employed null space projection methods, whereas zero acceleration constraint are enforced on feet ends in \cite{hereid_3d_2016}. The issue still remains that these methods are extremely computationally expensive and cannot be implemented in real time, taking a few minutes to solve the TO problem. For works that use reduced order models such as the centroidal dynamics or utilize zero moment point (ZMP) based methods as in \cite{pratt2006capture} and \cite{winkler_planning_2015}, experimental results of online-optimization are available however they are restricted to pseudo-static gaits rather than dynamic ones.


\noindent ii) In optimization-based control schemes, the goal is to compute constraint-aware feedback stabilizing control loops. This is most commonly achieved through a predictive framework, usually by creating a linearized model over finite time horizon. For instance in \cite{bledt2018cheetah}, \cite{fahmi_passive_2019-2}, \cite{angelini2019online} reduced-order model around the center of mass are used in a hierarchical framework to generate reference acceleration for the lower level controller to track. The downfall of these options are the need to linearize and/or simplify the underlying dynamics of the robot in order to make it feasible in real-time, and as a result not all constraints can be accounted for.
In \cite{hutter2014quadrupedal}, \cite{xin2020optimization}, \cite{galloway2015torque}, \cite{nguyen2016exponential} the desired control inputs are computed taking into account the full dynamics, and then optimization is carried out on a simplified least square or QP problem for tractability.

\noindent iii) A different approach is to remove optimization from the control strategy and instead modify the reference trajectories to obey desired constraints. This idea was popularized through reference governors (RG) \cite{kapasouris1988design}, where an efficient online optimization method is employed. The core of the idea being that the reference trajectory that the controller must follow can be manipulated while keeping it close to the original trajectory in the event that boundaries created by constraints are to be violated. Since its inception this idea has spawned many variations \cite{kolmanovsky2012developments}, \cite{garone2017reference} including an optimization free approach known as explicit reference governor (ERG) \cite{nicotra2015explicit}. Besides the possibility of utilizing an optimization-free form, the other major advantage with RG is that it acts as a add-on scheme to an existing controller without the need of any modification on the control scheme. 

While gait re-design in a reactive fashion is widely used in quasi-static walkers, one major reason that within-strides gait adjustment is less common in dynamic walkers is that their small support polygons, which can be as small as a point contact, leave small to no stability margins to avoid fall-overs. With this observation, we aim to apply a new control action, i.e., thrusters, during each gait cycle. 

We will employ thruster actions during a small part of a gait cycle to secure hybrid invariance. In this paper we extend our previous works \cite{dangol2020performance,dangol2020towards} by offering a systematic method, and modify the RG formulation to work with a popular framework known as hybrid zero dynamics (HZD) used to create stable gaits in bipeds.

\section{Overview of Control Design Idea}

In this section, we outline an overview of our approach to satisfy state, control and GRF constraints during gait cycles by \textit{deforming} the Zero Dynamics Manifolds describing the gaits which will be described in brevity here. 

The governing equations of motion are derived using the methods of Lagrange by taking the kinetic and potential energies of the system. The resulting equations of motion are expressed in vector form as following 
\begin{equation} 
    D_s(q_s)\ddot{q}_s + H_s(q_s,\dot{q}_s) = B_s(q_s)u,
\label{eq:full-model}    
\end{equation}

\begin{figure*}[t]
    \centering
    \includegraphics[width=1\linewidth]{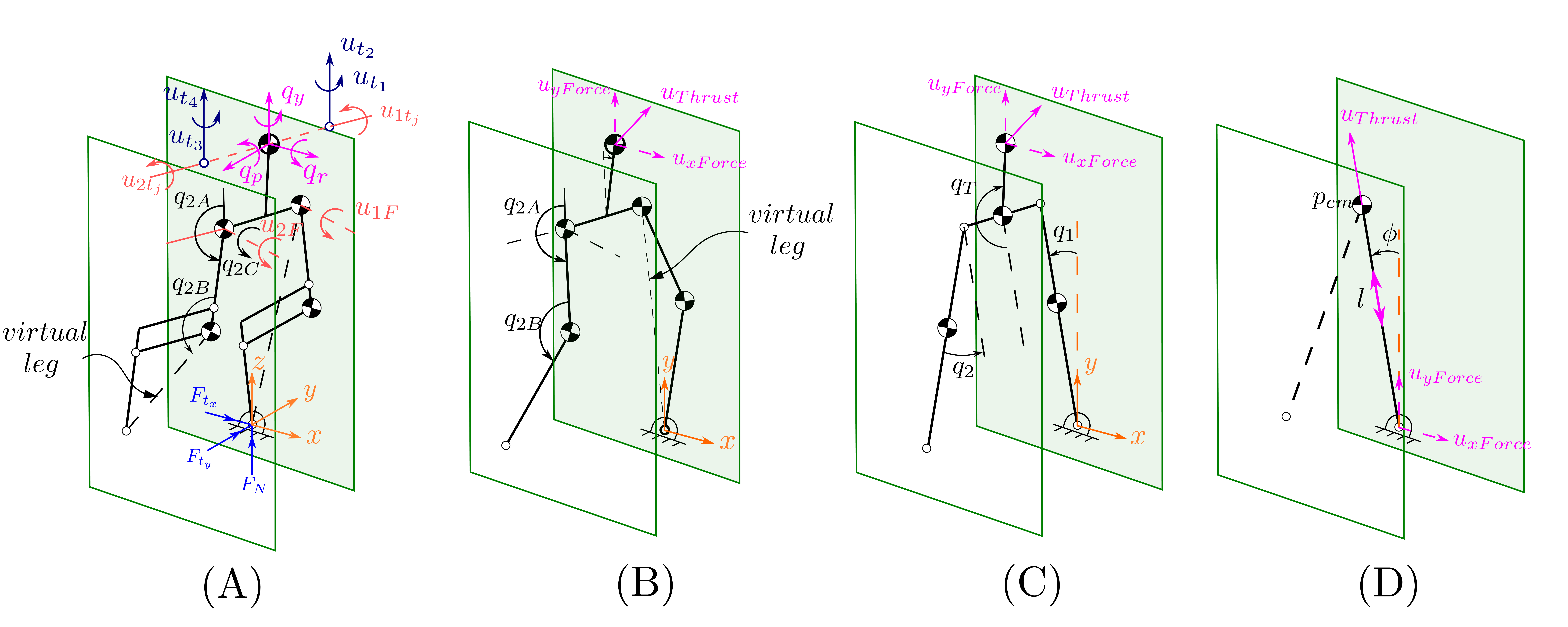}
    \vspace{-17pt}
    \caption{A) Stick figure diagrams which represents the thruster-assisted legged model, here the green boxes are the planes where each legs lies in. B,C,D) reduced order models, starting from 5-link, 3-link and VLIP models respectively.}
    \vspace{-14pt}
    \label{fig:stick-fig}
\end{figure*}

\noindent where $q_s$ is the configuration variable vector, $D_s(q_s)$ is the symmetric inertial matrix and is only dependent on $q_s$, the $H_s(q_s,\dot{q}_s)$ matrix contains Coriolis and gravity terms, and the control matrix $B_s(q_s)$ maps the inputs to the generalized coordinate accelerations $\ddot{q}_s$. Consequently, the full model can be written in state-space form as
\begin{equation}
    \dot{x}_s = \begin{bmatrix} 
    \dot{q}_s \\
    D_s^{-1}(-H_s + B_s u)
    \end{bmatrix}
    = f_s(x_s) + g_s(x_s)u,
\label{eq:ss-model}
\end{equation}
\noindent where the state vector is denoted by $x_s = [q^\top_s, \dot{q}^\top_s]^\top$. The notion of directional derivatives $L_{f}y = \frac{\partial y}{\partial x}f$ and holonomic constraints $y=h_s(x_s)=h_s(q_s)$ will be adopted in a similar way as mainstream publications in this field. The holonomic constraints ($y$) are widely known as Virtual Constraints (VCs) since they are enforced by closed-loop feedback. Based on these constraints celebrated control invariant sets of the form $\Gamma=\{[q_s^\top,\dot{q}_s^\top]^\top\in\mathbb{R}^{2n}|F(q_s,\dot{q}_s)=0\}$ where $F=[h_s(x_s)^\top,L_{f_s}h_s(x_s)^\top]^\top$ and $rank\{L_{f_s}h_s(x_s)\}=n-1$ can be defined. The rank condition guarantees that $\Gamma$, i.e., $q_s=h_s^{-1}(0)$, is $1$-dimensional.  Since there is only one degree of under-actuation (DOU) in our model, therefore, $h_s^{-1}(0)$ takes the form of a closed curve. We consider the following parametric descriptions  $q_1=r_1(q_n),\dots,q_{n-1}=r_{n-1}(q_n)$ where $q_n$ in our planar model is the last entry of $q_s$. As a result, the output function takes the following form:
\begin{equation}
h_s(q_s)=Hq_s-r(q_n)=q_{Act}-r(q_n)
\label{eq:vc}
\end{equation}
\noindent Note that $r=[r_1,\dots,r_{n-1}]^\top$ and the matrix $H\in\mathbb{R}^{(n-1)\times n}$ can take a trivial form if each joint independently is derived with a single actuator only, here $q_{Act}$ denotes the actuated variables. 

In a nutshell, the idea is to continuously deform $q_s=h_s^{-1}(0)$ such that the following conditions are satisfied. First, we want $h^{-1}_s(0)$ to remain continuous and closed curves, i.e., $r(q_n(t))=r(q_n(t+T))$ where $T$ is the gait period. Second, we want $\Gamma$ remain stabilizable at all times otherwise enforcing VCs will be impossible. Last, we want gait feasibility conditions including equality $C_{eq}(q_n,\dot{q}_n)=0$ and inequality constraints $0\leq C_{ineq}(q_n,\dot{q}_n)$ to be satisfied. This problem can take the following constrained ordinary differential equation form:
\begin{equation}
\left\{\begin{array}{ll}
\ddot{q}_n&=a_1(q_n) + a_2(q_n)\dot{q}^2_n + \mathrm{effects~of~deforming}~h^{-1}_s(0) \\
0 &\neq \frac{\partial h_s}{\partial x_s}\left[0^\top,\left[D_s^{-1}(q_n)B_s(q_n)\right]^\top\right]^\top ~ \mathrm{for~deformed~ h^{-1}_s(0)}\\
0&=C_{eq}(q_n,\dot{q}_n)\\
0&\leq C_{ineq}(q_n,\dot{q}_n)
\end{array}\right.
\label{eq:c-ode}
\end{equation}

\noindent where, the first line governs restriction dynamics and the second line is the condition for the stabilizability of $\Gamma$. Widely considered gait feasibility constraints such as $q_n(t+T)=q_n(t)$, $\dot{q}_n(t+T)=\dot{q}_n(t)$, $|x_s|<x_{m a x}$, $|u|<u_{m a x}$, $0<F_{N}$ and $|\frac{F_{\tau}}{F_{N}}|<\mu$ can potentially form the equality and inequality constraints, where $F_{\tau},F_{N}$ are the tangential and normal GRF respectively. 

This problem can be looked upon as a classical time-invariant, trajectory tracking problem, i.e., $q_{Act}$ parameterized in terms of $q_n$.  It can be easily resolved using optimization. In our approach, to solve the problem in an optimization-free fashion (or minimally use optimization) positive invariance property of $\Gamma$ -- i.e., being able to find the control input $u$ such that particularly when $q_s(0)$ is on $q_s=h_s^{-1}(0)$, $\dot{q}_s(t)$ remains tangent to $q_s=h^{-1}_s(0)$ yielding the solutions $q_s(t)$ remain on $\Gamma$ for all $t>0$ -- plays a key role and is closely dependent on how $q_s=h_s^{-1}(0)$ is deformed. 

In the case that this property is guaranteed, the constraint satisfaction problem can be transformed into a motion planning problem in the state space of the internal dynamics which can be conveniently tackled using simple path planning tools. This particularly becomes important and handy when fast and reactive gait planing is in need in dynamic walkers. We will further elaborate this concept with a simple example.

\section{Valid Deformations of  $q_s=h^{-1}_s(0)$}

Consider the state-space representation of the system dynamics given by ~\eqref{eq:ss-model}. Here, we will explore our options and choices in order to manipulate ~\eqref{eq:vc}. Our possible options are: i) scaling, i.e., $w_iq_n$, and ii) shifting, i.e., $q_n + w_i$, of the equilibrium point of the error function $y(q_s)$.

Evidently, constant priming terms yield a discrete collections of parameterized systems, i.e., $x_s=f_s(x_s)+g_s(x_s)u_{\omega}$, which is not desired here. Considering the numerical parameters $\omega$, which are used to parametereize $r(q_n,\omega)$, as auxiliary control input in discrete maps (e.g., Poincare maps) can only provide discontinuous means of priming $\Gamma_\omega$ at the boundaries. The continuous manipulation of these parameters can violate the transversality condition. Consider $y=q_{Act}-r(q_n,\omega)$. For a fixed $q_n$, it is possible to show that 
\begin{equation}
B_s^{*}(q_n)D_s(q_n)\left[r'(\omega(t),q_n)^\top,1\right]^\top
\label{eq:transversality}
\end{equation}
\noindent can vanish on $\Gamma_\omega$. Here, $B_s^{*}(q_n)=[0^{1\times(n-1)},1]$ and the annihilation of ~\eqref{eq:transversality} implies $g_s(x_s)$ and $\frac{\partial h_s(x_s)}{\partial x_s}$ are orthogonal. 

On the other hand, it is straightforward to show that $\dot{y}=\dot{q}_{Act}-r'(q_n)\dot{q}_n+\omega(t)$ yields relative degree $\left\{2,\dots,2\right\}$ on $\Gamma_\omega$. It is also noticeable that this choice of manipulating $y$ has no effects on $null\left\{\frac{\partial h_s}{\partial q_s}\right\}=\left[r'(q_n)^\top,1\right]$ which means that at least the primer has no influence over the transversality condition as long as $y=q_{Act}-r(q_n)$ is itself is valid VC.  

Notice that the role of the primer $\omega(t)$ in this form is comparable to the role of a disturbance term in the system given below
\begin{equation}
\left[\begin{array}{c}
\dot{y} \\
\ddot{y}
\end{array}\right] = 
\left[\begin{array}{l}
\dot{q}_{Act}-r'(q_n)\dot{q}_n \\
\gamma_1
\end{array}\right]
+
\left[\begin{array}{l}
0^{n-1} \\
\gamma_2
\end{array}\right]u+B_\omega \omega(t)
\label{eq:output-dyn}
\end{equation}

\noindent where $\gamma_1$ contains the feed-forward terms \cite{westervelt2018feedback}, $\gamma_2=-\left[I^{n-1},r'(q_n)\right]D^{-1}_s(q_s)B_s$ and $B_\omega=[I^{n-1},0^{n-1}]^\top$. As a result, the closed-loop system has to possess strong disturbance rejection properties. 

The roles of this disturbance term will be beneficial for us though as it will adjust the equilibria of $h_s(x_s)$ and $L_{f_s}h_s(x_s)$ under stabilising controllers with adjustable (and measurable) basins of attraction to successfully satisfy state, control and  GRF constraints.  To do this, we need to design an update law for the primer variable vector $\omega(t)$ such that the finite-time convergence of the solutions to $\Gamma_\omega$ in the closed-loop system is unaffected. 

As far as the design of $u$ is of concern, any nonlinear controller (or linear controllers if the nonlinear terms are bounded and the bounds are known) can satisfy the VCs. We will limit ourselves to the following modest feedback law $u=-L_g L_f h_s(x_s) ( L_f^2 h_s(x_s) + K_p y + K_d \dot{y} )$ where $K_i\in \mathbb{R}^{(n-1)\times(n-1)}$ are constant matrices and instead will remain focused on deforming $\Gamma_\omega$ in order to satisfy our constraints. Hence, we will assume stabilising supervisory controllers that guarantee the enforcement of the virtual constraints, however, their disturbance rejection properties has to be carefully considered. 

The control law given above can generate GAS at the equilibrium point of the system given by ~\eqref{eq:output-dyn} when the primer variable vector is time-invariant, i.e., $\omega(t)=\omega$.  Subsequently, $\Gamma_{\omega}$ takes the following form 
\begin{equation}
\Gamma_\omega=\left\{x_s \in X \mid h_s\left(x_{s}\right)=-K_P^{-1} K_{D} \omega(t), L_{f_{s}} h_{s}\left(x_{s}\right)=-\omega(t)\right\}
\label{eq:deformed-manifold}
\end{equation}

\noindent where the equilibrium points for $h_s(x_s)$ and $L_{f_s}h_s(x_s)$ under $\omega(t)$ are obtained by solving 
\begin{equation}
\left[\begin{array}{l}
y_\omega\\
\dot{y}_\omega
\end{array}\right]=\left[\begin{array}{cc}
-K_{P}^{-1} K_{D} & K_{P}^{-1} \\
-I^{n-1} & 0
\end{array}\right]\left[\begin{array}{l}
\omega(t) \\
0
\end{array}\right]
\label{eq:ss_sol}
\end{equation}

\noindent Of course, realizing GAS property under $\omega(t)$, i.e., when the primer variable is time-varying will not be achieved in a trivial fashion and requires an extra condition to be satisfied. We will discuss this in the proof of GAS property of the closed-loop system later. 

\section{Constraints Derivation}
\label{sec:constraint}

Consider the configuration variable vector $q_s=\left[q^\top_{Act},p^\top_1,q_n\right]^\top$ where $q_{Act}\in \mathbb{R}^m$ is the actuated joint angle, $m$ is the number of actuated joints and $p_1\in \mathbb{R}^2$ is the stance leg contact point. The control matrix $B_s(q_s)$ for $u=[u^\top_{Act},u^\top_{Thrust},F^\top]^\top$ in the Euler-Lagrange equations can take the following form
\begin{equation}
B^\top_s(q_s)=
\left[\begin{array}{ccc}
I^m & 0 & 0 \\
\frac{\partial p_{Thrust}}{\partial q_{Act}} & \frac{\partial p_{Thrust}}{\partial p_{1}} & \frac{\partial p_{Thrust}}{{\partial q_n}} \\
0 & I^2 & 0 
\end{array}\right]
\label{eq:ctrl-matrix}
\end{equation}

\noindent where $p_{Thrust}$ is the physical location of the thruster action $u_{Thrust}$. Notice that based on how the thruster actions look like the underactuated coordinate $q_n$ can be actuated. The following restriction dynamics can be obtained at every point on $\Gamma_\omega$ 
\begin{equation}
\ddot{q}_n=-\alpha^{-1}\left(\beta_1 \left(\dot{q}_n\right)^2+\beta_{2}\right)
\label{eq:rest-dyn}
\end{equation}

\noindent In this equation $\beta_2 = B^*_s G_s(q_n)$, where $G_s(q_n)$ contains terms affected by gravity, and $\beta_1$ is given by
\begin{equation}
\beta_1 = B^*_s\left(D_s(q_n)\left[r''^\top,0\right]^\top+\Sigma^n_{i=1}\left[r'^\top,1\right]^\top Q_i(q_n) \left[r'^\top,1\right]\right)
\label{eq:beta1}
\end{equation}

\noindent where $Q_i(q_n)$ are the Christoffel Symbols. A similar algebraic relationship for the constraints $[u^\top_{Act},F^\top]^\top$ can be obtained which is skipped here. 

Next, we will steer $y$ and $\dot{y}$ using the primer variable $\omega(t)$ in ~\eqref{eq:output-dyn} in order to make sure the solutions of ~\eqref{eq:rest-dyn} stay within the constraint-admissible space. To do this, consider the $y$-$\dot{y}$ space. 

Since we assumed a pre-stabilized system -- in fact all of the above derivations only make sense if $q_{Act}=r(q_n)+\int_0^\tau\omega(\tau)d\tau$ and $\dot{q}_{Act}=r'(q_n)\dot{q}_n+\omega(t)$ -- it is reasonable to evaluate the constraints $c_l\leq[u^\top_{Act},F^\top]^\top\leq c_u$ ($c_l$ and $c_u$ are constraint lower and upper bounds) based on the steady-state solutions, i.e., $y_\omega$ and $\dot{y}_\omega$, and ignore the transient solutions. 

Other than simplifying the nonlinear constraint satisfaction problem given in ~\eqref{eq:c-ode}, considering $[y^\top_\omega,\dot{y}^\top_\omega]^\top$ has another interesting result which will be explained below. Consider the set 
\begin{equation}
Y_{\omega}=\left\{z=[y^\top,\dot{y}^\top]^\top |[I^{n-1},-K^{-1}_PK_D]z=0\right\}
\label{eq:ss-rep-set}
\end{equation}

\noindent which is the locus of all of the steady-state solutions of the system ~\eqref{eq:output-dyn}.  It is possible to show that invariant sets around any point $[y^\top_\omega,\dot{y}^\top_\omega]^\top$ in the set defined by $Y_\omega$ can be created.

\section{Simulation Results and Concluding Remarks}

Fig.~\ref{fig:stick-fig} shows the low-order representations of our legged robot with two thrusters attached to its torso. This section discusses the simulation setup and results where the proposed framework is implemented on an equivalent low-order model (e.g., variable-length inverted pendulum model) of the full system. In this way, the low-order model is considered as a template used to satisfy the GRF, state, etc., constraints and the full system's joint angles and velocities are resolved through the forward kinematics equations. The simulation is done to show that the constraints, which are highly nonlinear and are often resolved using costly optimizers, can be enforced in a completely optimization-free fashion. 

\begin{figure}[t]
    \centering
    \includegraphics[width=1\linewidth]{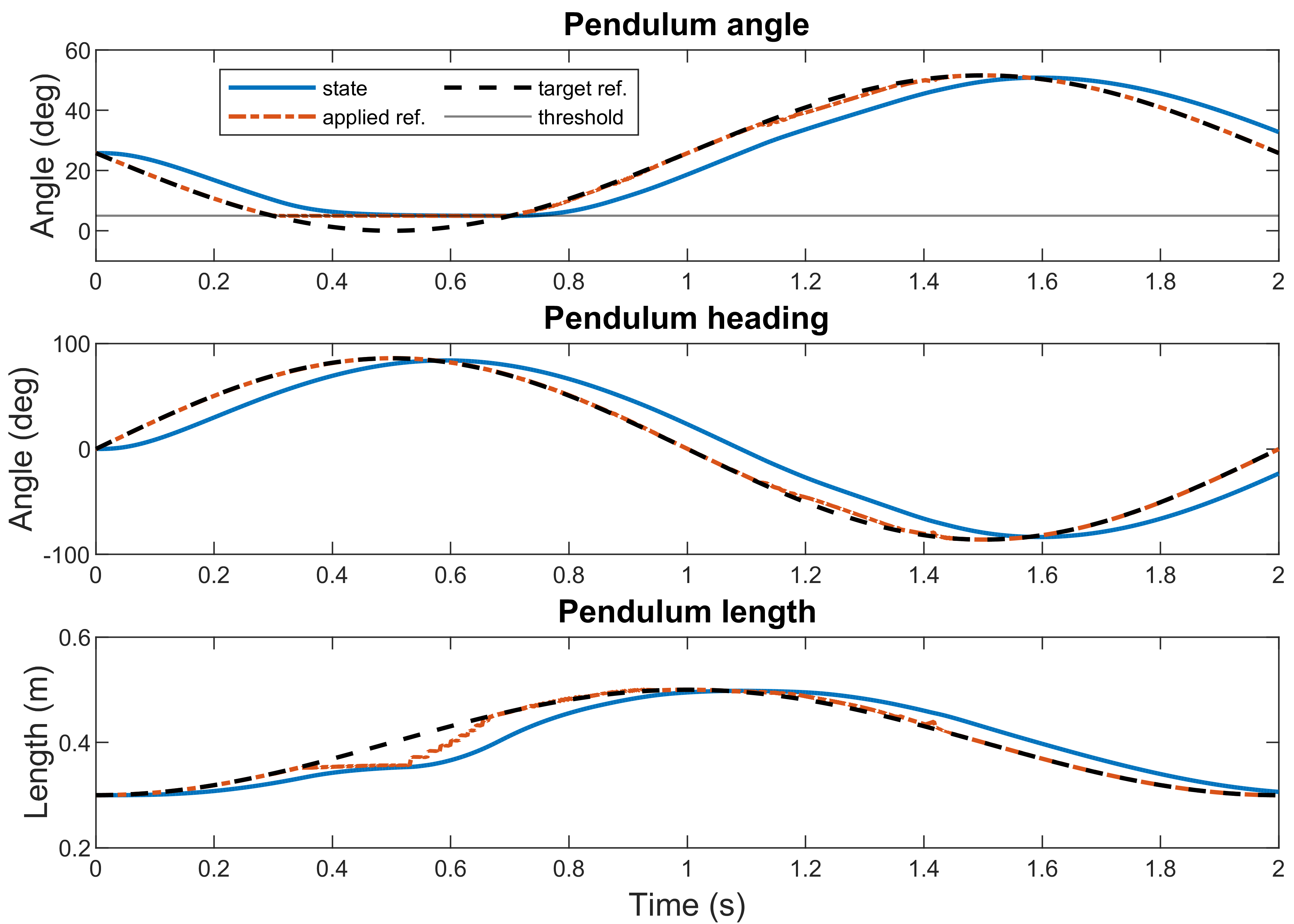}
    \vspace{-17pt}
    \caption{Illustration of the states constraint satisfaction. The primer update law was applied to adjust the joints reference trajectories to satisfy the constraints. This can be seen in the pendulum angle and virtual leg length, i.e., the low-order representation of the Harpy system, at 0.5 s. The algorithm prevents the pendulum angle from dropping to below 5$^\circ$ as specified by the constraints.}
    \label{fig:plot_vlip_states}
    \vspace{-10pt}
\end{figure}

The simulation has been resolved using a 4-th order Runge-Kutta scheme. The primer algorithm explained above was used to perform the trajectory manipulation task on the state trajectories in such a way that the constraints are satisfied. The following states in the low-order system were considered: the pendulum angle from the ground plane normal vector, heading angle, and the length of the virtual leg. Then, the following four constraints were stacked inside the nonlinear vector function $C_{ineq}$ in the inequality constraint equation (i.e., $C_{ineq} \geq 0$). The constraints include: a minimum pendulum angle of five degrees, the ground friction cone constraints in the $x$ and $y$ directions in the 3D model, and a minimum ground normal forces of 20 N. The latter constraint guarantees that the stance leg-end always stays on the ground surface.  

\begin{figure}[t]
    \centering
    \includegraphics[width=1\linewidth]{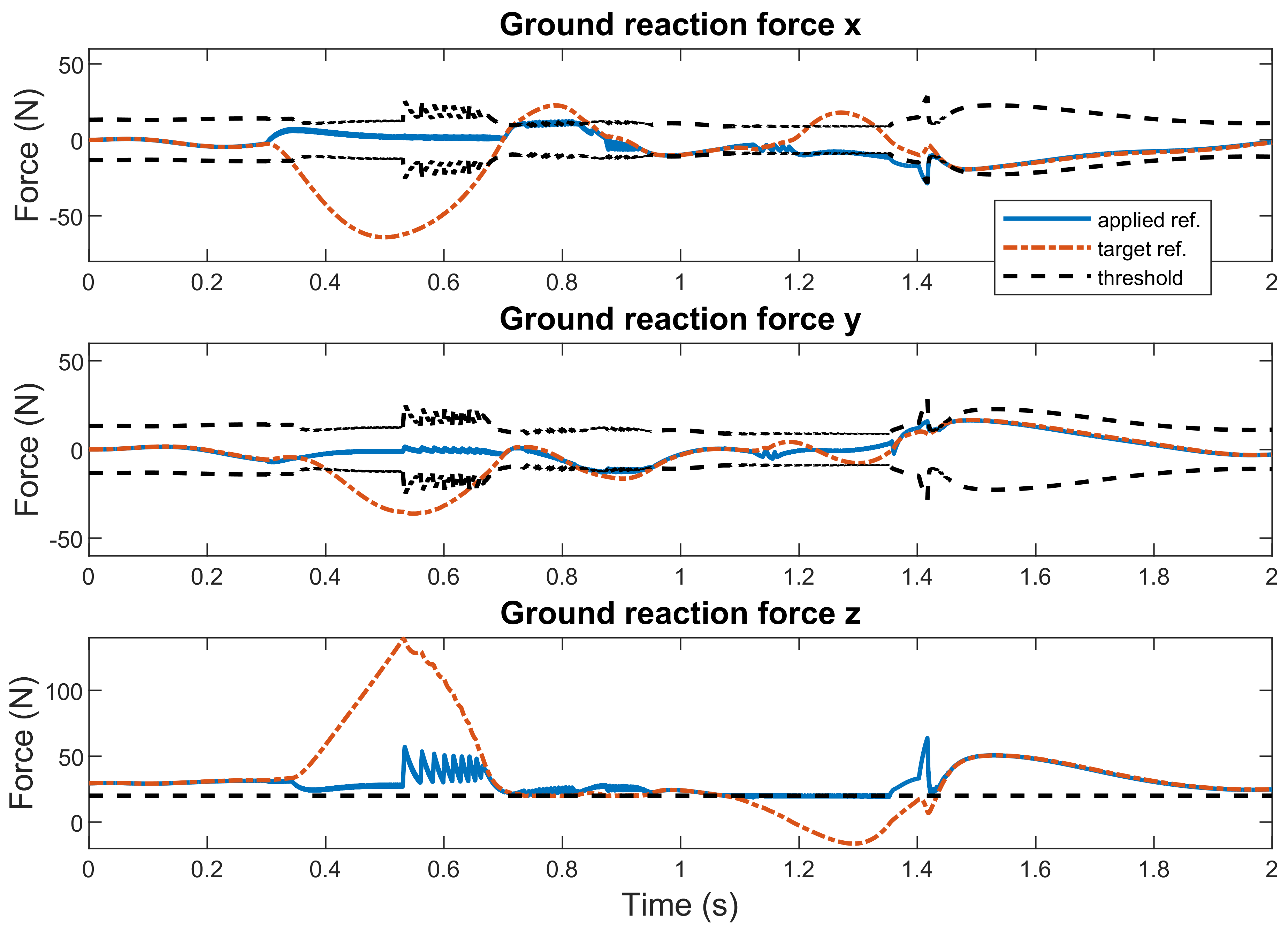}
    \vspace{-17pt}
    \caption{Illustration of the GRF constraints at the contact point (i.e., stance leg-end).}
    \label{fig:plot_vlip_grf}
    \vspace{-14pt}
\end{figure}

\begin{figure}[ht]
    \centering
    \includegraphics[width=1\linewidth]{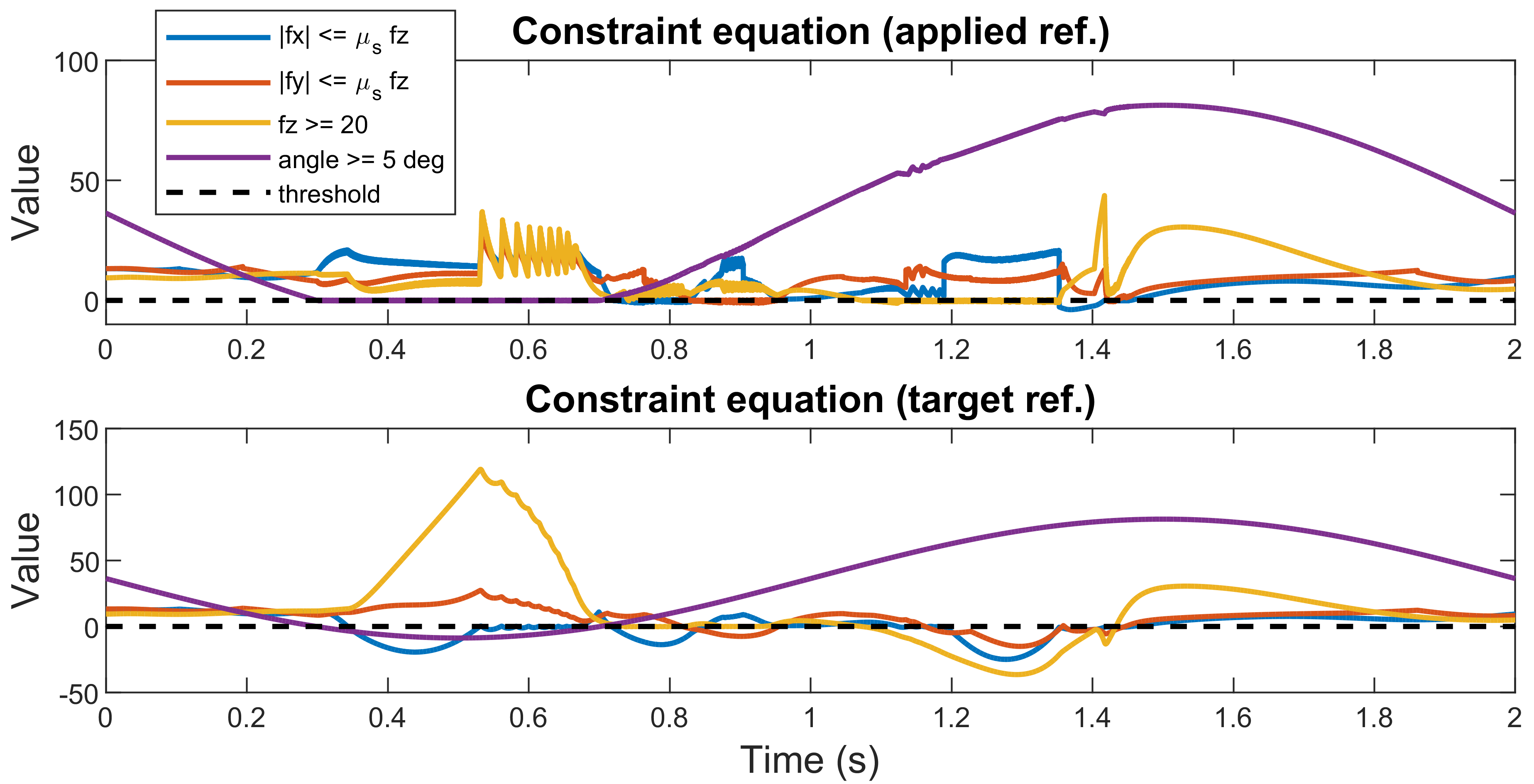}
    \vspace{-17pt}
    \caption{A comparison between all of the constraints involved when primed (top) and un-primed trajectories are applied. Almost all of the constraints are violated when the primer is absent.}
    \label{fig:plot_vlip_constraints}
    \vspace{-14pt}
\end{figure}

The simulation results are illustrated in Figs.~\ref{fig:plot_vlip_states} to \ref{fig:plot_vlip_constraints}. Figure \ref{fig:plot_vlip_states} displays the state trajectories for the low-order model. As it can be seen in this figure, a major deviation from the target reference at $t = 0.5$ s is required in order to avoid constraint violation. Fig. \ref{fig:plot_vlip_grf} shows the GRF profiles obtained using the target references and compares them against the primed references. While the target references frequently violate the no-slip constraints, the primed trajectories operate within the permissible bounds (this can be easily verified in Fig.~\ref{fig:plot_vlip_grf}). 

The constraints behavior under the proposed priming framework is summarized in Fig.~\ref{fig:plot_vlip_constraints}. This figure shows that the target reference lead to many violations while the primed references do not violate the constraints. However, in a few occasions, as it can be seen in Fig.~\ref{fig:plot_vlip_constraints}, the primed references violate the constraints. Based on our assumption outlined above in Section \ref{sec:constraint}, this temporary violations are expected. And, as it is evidently seen in, the trajectories are attracted to the constraint admissible sets after constraint violations occur.

\bibliographystyle{IEEEtran}
\bibliography{references,ref}

\end{document}